\documentclass[twocolumn]{article}
\usepackage[utf8]{inputenc}
\usepackage{microtype}
\usepackage{hyperref}
\usepackage{amssymb}
\usepackage{amsmath}
\usepackage{dsfont}
\usepackage{algorithm}
\usepackage{algorithmic}

\usepackage{amsthm}
\usepackage{thmtools, thm-restate}
\declaretheorem[numberwithin=section]{thm}
\declaretheorem[sibling=thm]{theorem}
\declaretheorem[sibling=thm]{lemma}

\newcommand{\ie}{i.e., }

\newcommand{\ind}{\mathds{1}}

\usepackage{amsmath,amssymb}
\usepackage{hyperref}
\usepackage{color}
\usepackage{amsbsy}
\usepackage{relsize}

\newtheorem{proposition}{Proposition}

\title{Statistical Analysis of Policy Space Compression Problem}
\author{ Majid Molaei, Marcello Restelli, Alberto Maria Metelli, Matteo Papini }
\date{}

\begin{document}

\maketitle

\begin{abstract}
Policy search methods are crucial in reinforcement learning, offering a framework to address continuous state-action and partially observable problems.
However, the complexity of exploring vast policy spaces can lead to significant inefficiencies. Reducing the policy space through policy compression emerges as a powerful, reward-free approach to accelerate the learning process. This technique condenses the policy space into a smaller, representative set while maintaining most of the original effectiveness. 
Our research focuses on determining the necessary sample size to learn this compressed set accurately. We employ Rényi divergence to measure the similarity between true and estimated policy distributions, establishing error bounds for good approximations. To simplify the analysis, we employ the $l_1$ norm, determining sample size requirements for both model-based and model-free settings. Finally, we correlate the error bounds from the $l_1$ norm with those from Rényi divergence, distinguishing between policies near the vertices and those in the middle of the policy space, to determine the lower and upper bounds for the required sample sizes.
\end{abstract}

\section{Introduction} \label{S:Introduction}

Over recent decades, reinforcement learning \cite{sutton2018reinforcement} has become a powerful tool for tackling sequential decision-making under uncertainty, with notable successes in areas such as game-solving, robotic manipulation, and text generation. However, achieving these breakthroughs often requires reinforcement learning algorithms to be trained on a massive number of samples collected from the environment, limiting their applicability to a broader range of real-world problems. A significant contributor to this sample inefficiency is the vast size of the so-called policy space—the set of decision strategies from which the optimal one is learned. The more decision strategies available, the greater the potential for achieving desirable performance, but this also makes the learning process less efficient.

A key approach in reinforcement learning (RL) for solving complex decision-making problems is the use of policy search methods \cite{peters2008reinforcement}. These methods involve defining a parametric policy space, where the policy is represented as a function parameterized by a set of variables. By optimizing these parameters, policy search aims to discover a policy that maximizes the expected reward in a given environment. This approach allows for flexibility in representing a wide range of policies, making it particularly effective for handling high-dimensional and continuous action spaces, as well as for tasks that require sophisticated strategies. Notable methods in this area include policy gradient techniques like REINFORCE \cite{williams1992simple} and trust region policy optimization (TRPO) \cite{pmlr-v37-schulman15}, which have been widely used in both discrete and continuous action spaces.

Recently, online learning methods have been proposed that perform policy search over a set of finite stochastic policies, providing theoretical guarantees about regret minimization \cite{papini2019optimistic,metelli2021policy}. However, the effectiveness of these approaches depends heavily on how well the set of candidate policies covers the broader policy space. A poor representation of the policy space can lead to suboptimal performance, as the policy search may overlook better strategies. Thus, constructing a well-distributed and diverse set of policies is crucial for the success of these methods.

A recent study \cite{mutti2022reward} addresses this problem by proposing a pre-processing step to compress the set of decision strategies. This process, known as policy space compression, aims to reduce the size of the policy space while retaining most of its expressive power. The idea is that the risk of a slight reduction in performance due to the reduction of the policy space is compensated by improved efficiency. In practice, policy space compression seeks to identify a small set of $K$ representative policies that effectively cover the state-action distributions induced by all policies in the original policy space, with the coverage being approximated by a threshold $\sigma$. While an algorithm for policy space compression is also presented in \cite{mutti2022reward}, some important questions remain unanswered, especially from a statistical point of view.

The main goal of our research is to statistically address the problem and, consequently, determine the number of sampled interactions necessary to learn an approximate compression of policies in a Markov Decision Process (MDP) \cite{puterman1994markov}, ensuring with high confidence that the compression is nearly optimal. To achieve this goal, we will undertake several steps: first, we will utilize an optimization oracle to compute this compression function on estimated MDPs, allowing us to separate the statistical analysis from computational aspects. We then, in Section \ref{problem_formulation}, formalize the statistical problem using Rényi divergence \cite{renyi1961measures}, which measures the similarity between the exact and estimated state-action pair distributions of policies. Consequently, in Section \ref{error_range}, we study the meaningful error range because, beyond a specific threshold, the estimated distribution of state-action pairs is guaranteed to be close to the true distribution without needing additional samples. Then, in Section\ref{total_variation}, we reformulate our original statistical problem by transitioning from Rényi divergence to the Total Variation divergence to simplify our statistical analysis. We determine the number of samples needed for both model-based and model-free settings \cite{sutton2018reinforcement}. Subsequently, in Section \ref{Analysis_with_Rényi_Divergence}, we return to our original problem by identifying a correlation between error bounds from the Total Variation divergence and those from Rényi divergence between policies, thus determining the lower and upper bounds for the number of samples required according to our original statistical formulation.

\section{Preliminaries}\label{S:Preliminaries}

In this section, we introduce essential concepts and mathematical formulations necessary for understanding the subsequent sections of this paper. These include controlled Markov processes, policy optimization, importance sampling, and Rényi divergence.

\subsection{Controlled Markov Processes}

A \textit{Controlled Markov Process (CMP)} \cite{puterman1994markov} is a mathematical model used to describe a system that transitions between states based on specific actions. Formally, it is defined as a tuple:
\begin{equation*}
    \mathcal{M} := (S, A, P, \mu, \gamma).
\end{equation*}
where:
\begin{itemize}
    \item $S$: The set of all possible states of the system.
    \item $A$: The set of all possible actions that an agent can take.
    \item $P: S \times A \to \Delta(S)$: The transition model, where $P(s'|s, a)$ represents the probability of transitioning to state $s'$ from state $s$ when action $a$ is taken.
    \item $\mu: \Delta(S)$: The initial state distribution, indicating the probability distribution over states at the beginning of the process.
    \item $\gamma \in (0, 1]$: The discount factor, which models the preference for immediate rewards over future rewards.
\end{itemize}

An agent interacts with the CMP through a policy $\pi_\theta: S \to \Delta(A)$, where $\pi_\theta(a|s)$ represents the probability of taking action $a$ in state $s$. The policy is often parameterized by $\theta$, which can be adjusted to optimize performance.

The \textit{$\gamma$-discounted state distribution} $d_\theta(s)$, which represents the distribution of states over time under policy $\pi_\theta$, is given by:
\begin{equation*}
    d_\theta(s) = (1 - \gamma) \sum_{t=0}^{\infty} \gamma^t \Pr(s_t = s | \pi_\theta).
\end{equation*}
This can also be expressed recursively as:
    \begin{align*}
    &d_\theta(s) = (1 - \gamma) \mu(s) \\
    &\quad \quad \quad \quad + \gamma \sum_{s'\in S}\sum_{a' \in A} d_s^{\pi_\theta}(s') \pi_\theta(a'|s') P(s|s', a').
    \end{align*}

The \textit{$\gamma$-discounted state-action distribution} $d_\theta(s, a)$, which represents the joint distribution over states and actions under policy $\pi_\theta$, is defined as:
\begin{equation*}
    d_\theta(s, a) = \pi_\theta(a|s) d_\theta(s).
\end{equation*}

\subsection{Policy Optimization}

\textit{Policy Optimization (PO)} is a central concept in reinforcement learning, focusing on finding the policy $\pi_\theta$ that maximizes the expected cumulative reward. The performance of a policy is defined by the objective function:
\begin{equation*}
    J(\theta) := \mathbb{E} \left[ \sum_{t=0}^{\infty} \gamma^t R(s_t, a_t) \right] = \frac{1}{1 - \gamma} \mathbb{E}_{(s, a) \sim d_\theta} [R(s, a)].
\end{equation*}
where $R(s, a)$ is the reward received after taking action $a$ in state $s$. This expectation is taken over the state-action distribution induced by the policy $\pi_\theta$.

To estimate $J(\theta)$, we can use a Monte Carlo method with a batch of $N$ samples $\{s_n, a_n\}_{n=1}^N$ generated by policy $\pi_\theta$:
\begin{equation*}
    \hat{J}(\theta) = \frac{1}{(1 - \gamma) N} \sum_{n=1}^N R(s_n, a_n).
\end{equation*}

\subsection{Importance Sampling and Rényi Divergence}

\textit{Importance Sampling (IS)} \cite{gilks1995tutorial} \cite{thomas2002importance} is a technique used to estimate properties of a particular distribution while only having samples generated from a different distribution. This is particularly useful in policy optimization, where we might want to evaluate a target policy $\pi_{\theta'}$ using samples from a behavior policy $\pi_\theta$. The importance weight, which re-weights the samples from the behavior policy to represent the target policy, is given by:
\begin{equation*}
    w_{\theta'/\theta}(s, a) := \frac{d_{\theta'}(s, a)}{d_\theta(s, a)}.
\end{equation*}

Using these weights, a Monte Carlo estimate of the performance of the target policy $\pi_{\theta'}$ can be expressed as:
\begin{equation*}
    \hat{J}_{\text{IS}}(\theta'/\theta) = \frac{1}{(1 - \gamma) N} \sum_{n=1}^N w_{\theta'/\theta}(s_n, a_n) R(s_n, a_n).
\end{equation*}

The variance of the importance weights is related to the \textit{exponentiated 2-Rényi divergence} $D_2$, a measure of similarity between two distributions. The variance can be expressed as:
\begin{equation*}
    \text{Var}_{(s, a) \sim d_{\theta}}[w_{\theta'/\theta}(s, a)] = D_2\big(d_{\theta'} \| d_{\theta}\big) - 1.
\end{equation*}
where the exponentiated 2-Rényi divergence is defined as:
\begin{equation*}
    D_2\big(d_{\theta'} \| d_{\theta}\big) := \sum_{s \in S}\sum_{a \in A} d_{\theta}(s, a)\left( \frac{d_{\theta'}(s, a)}{d_{\theta}(s, a)} \right)^2.
\end{equation*}

The variance of the IS estimator is bounded by:
\begin{equation*}
    \text{Var}_{(s, a) \sim d_{\theta}}[\hat{J}_{\text{IS}}(\theta'/\theta)] \le \left( \frac{R_{\max}}{1 - \gamma} \right)^2 \frac{D_2\big(d_{\theta'} \| d_{\theta}\big)}{N}.
\end{equation*}

\subsection{Concentration Bounds for Empirical Estimates}

Before moving forward with the analysis, it is worth reporting two useful results on the Hoeffding-type concentration of the empirical estimates \cite{weissman2003inequalities} \cite{efroni2021reinforcement} \cite{leon2020optimal}.

\begin{lemma}\label{L:1}
    Let $\{ X_i \}_{i = 1}^N$ be i.i.d. random values over $[a]$ such that $\mathbb{P} (X_i = m) = p_m$, and let $\hat{p}_m = \frac{1}{N} \sum_{i = 1}^N \ind (X_i = m)$ be the empirical estimate of $p_m$. For every confidence $\delta \in (0, 1)$, it holds:
    \begin{equation*}
        \mathbb{P} \left( \| p_m - \hat{p}_m \|_1 \geq \sqrt{\frac{2a \ln \frac{2}{\delta}}{N}} \right) \leq \delta.
    \end{equation*}
    \label{thr:concentration_empirical_distribution}
\end{lemma}

\begin{lemma}\label{L:2}
    For a function \( f: [d] \to [0, 1] \) and a sample \( X_0 \sim \pi \) drawn from the stationary distribution, the following holds:    
\[
\mathbb{P}\left(\left| \frac{1}{N} \sum_{n=1}^{N} f(X_n) - \mu \right| \geq \epsilon \right) \leq 2 \exp \left\{ - \frac{\gamma_0}{2(2 - \gamma_0)} \epsilon^2 N \right\},
\]
where the eigenvalues of \( P \) are ordered in decreasing fashion as \( 1 > \lambda_2(P) \geq \cdots \geq \lambda_d(P) \), and the spectral gap of \( P \) is defined as:
\[
\gamma_0 := \min \{1 - \lambda_2(P), 1\}.
\]
    \label{thr:concentration_empirical_distribution}
\end{lemma}

Understanding these foundational concepts and their mathematical formulations is essential for the subsequent development of the policy space compression problem and associated algorithms. By leveraging these principles, we can develop more efficient and effective reinforcement learning methods.

\section{Problem Formulation} \label{problem_formulation}

\textit{Policy space compression} aims to identify a set of \( K \) representative policies whose corresponding state-action distributions are close enough (according to some divergence $D(\cdot||\cdot)$) to the state-action distribution $d_\theta$ of any policy $\pi_\theta$ in the original policy space. In particular, we want to guarantee that for every policy $\pi_\theta$ there exists at least one policy in the compressed set such that the divergence $D(\cdot||\cdot)$ between their state-action distributions is less than a threshold \( \sigma \):

\begin{align*}
    \max_{\theta \in \Theta} \min_{k \in [K]} D(d_\theta\|d_{\theta_k}) \leq \sigma.
\end{align*}

In the following, we will consider two divergences: 
\begin{itemize}
    \item Total variation
    $$D_{TV}(d_\theta\|d_{\theta_k}) = \frac{1}{2}\sum_{s\in S}\sum_{a \in A} \left|d_\theta(s,a)-d_{\theta_k}(s,a)\right| $$
    \item Exponentiated Réyni divergence of order $2$
    $$D_2\big(d_\theta \| d_{\theta_{k}}\big) = \sum_{s\in S}\sum_{a \in A} \frac{d^{2}_{\theta}(s,a)}{d_{\theta_{k}}(s,a)}.
 $$
\end{itemize}

In this paper, we assume access to an optimization oracle, allowing us to separate the statistical analysis from the computational aspects of the problem.
At this stage, our primary focus is to analyze the policy space compression problem from a statistical standpoint.

Derived from the optimization problem mentioned above, the statistical study of the compression problem can be defined as follows: Determining the number of samples needed to ensure, with a predefined probability \(\delta\), that each estimated vector of the state-action pair distributions induced by the \(k\) representative policies with parameters \(\theta_1\), \dots, \(\theta_k\) (i.e., \(\hat{d}_{\theta_1}\), \dots, \(\hat{d}_{\theta_k}\)), is within \(\sigma\) distance of the true vector (i.e., \(d_{\theta_1}\), \dots, \(d_{\theta_k}\)). All these vectors have \(|SA|\) components.

So, the problem can be formalized using the following inequality expression:

\begin{equation}
\mathbb{P} \left(D\big(\hat{d}_{\theta_{k}} \| d_{\theta_{k}}\big) > \sigma \right) \leq \delta,\;\;\;\;\;\;\;\;\; \forall k \in[K],
\nonumber
\end{equation}

where $\delta \in [0, 1]$ is a fixed confidence, and $\sigma$ is the threshold to the estimation error measured according to the selected divergence.

\section{Estimation Error Range} \label{error_range}

Let's first study the role of the estimation error threshold $\sigma$ in the statistical study of the policy space compression problem. First, we will revise the geometry of the problem. Then, we will discuss how to set the error threshold properly.

\subsection{Geometry of the Problem}

The policy space is a bounded convex polytope with $|SA|$ zero-dimensional vertices (the $i$-th vertex will be denoted by $(s,a)_i$), $\binom{|SA|}{2}$ one-dimensional edges, $\binom{|SA|}{3}$ two-dimensional faces, and so on, up to the original $|SA|$-dimensional polytope, which is essentially a shape similar to the convex hull of all the vertices.

The space of all state-action distributions is an $|SA|$-dimensional polytope, where the vertices represent the situation in which a single state is visited with probability one and a single action is taken in that state with probability one.

Having established the basic geometry, it is worth examining the values of the parameter $\sigma$, i.e., the estimation error threshold, from a geometric standpoint.

\subsection{Upper Bound to the Error Threshold}

It can be shown that $\sigma_{2} \geq |SA|$ or $\sigma_{TV} \geq \sqrt{\frac{|SA|-1}{|SA|}}$ is not meaningful, as demonstrated by the following proposition:

\begin{proposition}
If the threshold $\sigma_{2} \geq |SA|$ or $\sigma_{TV} \geq \sqrt{\frac{|SA|-1}{|SA|}}$, then even without a single sample, we can be assured that the estimated state-action pair distribution is within $\sigma$ distance of its exact value. This is because, in this case, with one representative policy—the policy that induces a uniform distribution over all state-action pairs—all the points inside the polytope space are within $\sigma$ distance of that representative policy.
\end{proposition}

\section{Analysis using Total Variation} \label{total_variation}

In this section, we provide lower bounds on the number of samples needed to ensure, in high probability ($1-\delta$), that the estimation error, measured using the total variation, is less than $\sigma_{TV}$:

\begin{equation}
\mathbb{P} \left(D_{TV}\big(\hat{d}_{\theta_{k}} \| d_{\theta_{k}}\big) > \sigma_{TV} \right) \leq \delta,\;\;\;\;\;\;\;\;\; \forall k \in[K].
\nonumber
\end{equation}

In particular, leveraging Lemmas \ref{L:1} and \ref{L:2}, we will provide results for scenarios where the transition model $P$ is either known or unknown.

\subsection{Number of Samples with Known Model}

In the case where the transition model $P$ is known, to determine the number of samples required to ensure that the estimated state-action pair distribution of a specific policy is within a $\sigma_{TV}$ distance of its exact value, we can apply Lemma \ref{L:2} directly:

\begin{align}
 &2 \exp \left\{ - \frac{\gamma_0}{8(2 - \gamma_0)} \sigma_{TV}^2 N \right\} = \delta
 \nonumber\\[8pt]
 &\Rightarrow N = \frac{8(2 - \gamma_0)}{\gamma_0 \sigma_{TV}^2} \ln \left( \frac{2}{\delta} \right).
 \nonumber
\end{align}

We should note that these samples can be collected by the agent interacting with the environment under our policy.

For $K$ policies, to ensure that the total variation distance between the exact and estimated values is at most $\sigma_{TV}$, the number of samples $N$ needed is:

\begin{equation}
N = \frac{2K(2 - \gamma_0)}{\gamma_0 \sigma_{TV}^2} \ln \left( \frac{2}{\delta} \right).
\nonumber
\end{equation}

\subsection{Number of Samples with Unknown Model}

In case the transition model $P$ is unknown, consider Lemma \ref{L:1} and also the following lemma:

\begin{lemma}
   The Total Variation divergence between the state-action distributions induced by $\pi$ over the true MDP, \ie, $d_{\theta_{k}}$, and the estimated MDP, \ie, $\hat{d}_{\theta_{k}}$, can be upper bounded as:
    \begin{equation*}
    D_{TV}\big(\hat{d}_{\theta_{k}} \| d_{\theta_{k}}\big) \leq \frac{\gamma}{1 - \gamma} \mathbb{E}_{(s, a) \sim d_{\theta_{k}}} \left[ \left\| \hat{P}(\cdot | s, a) - P (\cdot | s, a) \right\|_1 \right].
    \end{equation*}
    \label{thr:dist_upper_bound}
\end{lemma}

If, for each state-action pair among the $|SA|$ pairs, the Total Variation divergence between the exact and estimated transition probabilities across all states $s'$ is less than $\frac{2(1-\gamma)}{\gamma}\sigma_{TV}$, then the number of samples required is:

\begin{align}
 &\sqrt{\frac{2|S|\ln\left(\frac{2}{\delta}\right)}{N}} = \frac{2(1-\gamma)}{\gamma}\sigma_{TV}
 \nonumber\\[8pt]
 &\Rightarrow N = \frac{8\gamma^{2}|S|}{(1-\gamma)^{2}\sigma_{TV}^{2}}\ln\left(\frac{2}{\delta}\right).
 \nonumber
\end{align}

For all $|SA|$ state-action pairs, and consequently, for the final number of samples required for $K$ policies, we have:

\begin{equation}
N = \frac{8\gamma^{2}|S^{2}A|}{(1-\gamma)^{2}\sigma_{TV}^{2}}\ln\left(\frac{2}{\delta}\right).
\nonumber
\end{equation}

It is important to note that these samples can be obtained not by having the agent interact with the environment under our policy, but rather by using a generative model to produce samples of different state-action pairs.

\section{Analysis using Rényi Divergence} \label{Analysis_with_Rényi_Divergence}

In this section, we provide the sample complexity results in the case where the Rényi divergence is used to measure the distance between the state-action distributions induced by the policies.
\begin{equation}
\mathbb{P} \left(D_{2}\big(\hat{d}_{\theta_{k}} \| d_{\theta_{k}}\big) > \sigma_{2} \right) \leq \delta,\;\;\;\;\;\;\;\;\; \forall k \in[K].
\nonumber
\end{equation}

To exploit the results provided in the previous section, we need to establish a connection between the threshold considered for the Total Variation divergence ($\sigma_{TV}$) and the one for the Rényi divergence ($\sigma_2$).

\subsection{The Relationship between \texorpdfstring{$\sigma_2$}{sigma} and \texorpdfstring{$\sigma_{TV}$}{sigmaTV}}

In the following, we will derive upper and lower bounds to the threshold in Total Variation given the threshold in the 2-Rényi divergence.
First of all, we need to make some considerations about the threshold to be considered in the case of the Rényi divergence:

\begin{itemize}
    \item For a specific value of $\sigma_2$, the acceptable error range for our $K$ policies depends on the location of their state-action pair distribution vector within the polytope space. We can show that as we move towards the vertices of the polytope, the acceptable error range becomes tighter.
    \item For policies where the state-action pair distribution vector is near a vertex, the tightest acceptable error range is achieved when the probability value for the state-action pair associated with that vertex is fixed.
\end{itemize}

\subsubsection{Upper Bound to the Total Variation given the 2-Rényi divergence}

Here, we want to establish how large the Total Variation divergence between two distributions can be given their exponentiated 2-Rényi divergence.
To do this, we consider as a representative distribution $d_{\theta_{k}}$ the one with uniform probabilities over all the state-action pairs. In fact, to maximize the Total Variation divergence between two distributions subject to a constraint on the 2-Rényi divergence, one of the distributions needs to be the uniform one.  
To find a policy at the $\sigma_{2}$ distance from our representative policy $\theta_{k}$, we can use the following lemma:

\begin{lemma}
By moving away from the representative policy $\theta_{k}$ with uniformly distributed state-action pairs in the direction of a specific vertex, for instance $(s,a)_{1}$, at a $\sigma_{2}$ distance away from $\theta_{k}$, and considering:
\begin{align*}
&\hat{d}_{\theta_k}(s,a)_{2} = \hat{d}_{\theta_k}(s,a)_{3} = \dots = \hat{d}_{\theta_k}(s,a)_{|SA|} \\[8pt]
&=\frac{1 - \hat{d}_{\theta_k}(s,a)_{1}}{|SA|-1},
\end{align*}
we reach:
\begin{equation*}
\hat{d}_{\theta_k}(s,a)_{1} = \frac{1 \pm \sqrt{\big(|SA|-1\big)(\sigma_{2}-1)}}{|SA|},    
\end{equation*}
and:\\
\begin{align*}
& \hat{d}_{\theta_k}(s,a)_{2} = \hat{d}_{\theta_k}(s,a)_{3} = \dots = \hat{d}_{\theta_k}(s,a)_{|SA|} \\[8pt]
&= \frac{1 - \left(\frac{1 \pm \sqrt{\big(|SA|-1\big)(\sigma_{2}-1)}}{|SA|}\right)}{|SA|-1}.
\end{align*}
\end{lemma}

We are now ready to introduce an upper bound to the Total Variation divergence as a function of the 2-Rényi divergence.

\begin{theorem}
Given two state-action distributions with a 2-Rényi divergence equal to $\sigma_2$, their Total Variance divergence is upper bounded by $\max\sigma_{TV}$:
\begin{equation*}
\boldsymbol{\max\sigma_{TV} = \frac{\sqrt{(|SA|-1)(\sigma_{2}-1)}}{|SA|}}.
\end{equation*}
\end{theorem}

\subsubsection{Lower Bound to the Total Variation given the 2-Rényi divergence
}
To compute the minimum value that the Total Variation divergence can reach given a 2-Rényi divergence equal to $\sigma_2$, we consider as a representative state-action distribution $d_{\theta_{k}}$ one of those at $\sigma_2$ distance from a vertex, e.g., say $(s,a)_{1}$.
Since all these distributions have the same Total Variation with respect to the representative distribution, for simplicity, we consider the one that has equal probabilities for all the state-action pairs except for $(s,a)_{1}$.

The representative distribution can be determined by imposing the following constraint:

\begin{align*}
& \sum_{s\in S}\sum_{a \in A} \frac{\hat{d}^{2}_{\theta_k}(s,a)}{d_{\theta_{k}}(s,a)}\\[8pt]
&= \frac{(1)^{2}}{d_{\theta_{k}}(s,a)_{1}} + \frac{(0)^{2}}{d_{\theta_{k}}(s,a)_{2}} + \dots + \frac{(0)^{2}}{d_{\theta_{k}}(s,a)_{|SA|}} \\[8pt]
&= \sigma_{2}
\end{align*}

So, $d_{\theta_{k}}(s,a)_{1}$ will be equal to $\frac{1}{\sigma_{2}}$.
For the other state-action pairs, having assumed that they have the same probability

\begin{equation}
d_{\theta_{k}}(s,a)_{2} = d_{\theta_{k}}(s,a)_{3} = \dots = d_{\theta_{k}}(s,a)_{|SA|}
\nonumber
\end{equation}

and that

\begin{equation}
\sum_{s \in S}\sum_{a \in A} d_{\theta_{k}}(s,a) = 1,
\nonumber
\end{equation}

we have:

\begin{align*}
& \frac{1}{\sigma_{2}} + \big(|SA|-1\big)d_{\theta_{k}}(s,a)_{2} = 1 \\[8pt]
& \Rightarrow d_{\theta_{k}}(s,a)_{2} = d_{\theta_{k}}(s,a)_{3} = \dots = d_{\theta_{k}}(s,a)_{|SA|} \\[8pt]
& \quad \;\; = \frac{\sigma_{2}-1}{\sigma_{2}(|SA|-1)}
\end{align*}

As mentioned, to determine the acceptable error range, we need to identify a policy that is within $\sigma_{2}$ distance from our representative policy $\theta_{k}$. However, because we are not dealing with a spherical shape, policies that are at a $\sigma_{2}$ distance from $\theta_{k}$ are not necessarily at the same Euclidean distance from it.

For example, it is possible to find a policy that is at a $\sigma_{2}$ distance and simultaneously at the maximum Euclidean distance from our representative policy $\theta_{k}$.

To find such a policy, we can use the following lemma:

\begin{lemma}
By moving away from the representative policy $\theta_{k}$ in the direction of the specific vertex $(s,a)_{1}$, at a $\sigma_{2}$ distance from the representative policy $\theta_{k}$:
\begin{align*}
& d_{\theta_{k}}(s,a)_{2} = d_{\theta_{k}}(s,a)_{3} = \dots = d_{\theta_{k}}(s,a)_{|SA|} \\[8pt]
& = \frac{\sigma_{2}-1}{\sigma_{2}(|SA|-1)}
\end{align*}
and considering:
\begin{align*}
&\hat{d}_{\theta_k}(s,a)_{2} = \hat{d}_{\theta_k}(s,a)_{3} = \dots = \hat{d}_{\theta_k}(s,a)_{|SA|} \\[8pt]
&= \frac{1 - \hat{d}_{\theta_k}(s,a)_{1}}{|SA|-1},
\end{align*}
we reach:
\begin{equation*}
\hat{d}_{\theta_k}(s,a)_{1} = 1 \quad \text{or} \quad \hat{d}_{\theta_k}(s,a)_{1} = \frac{2-\sigma_{2}}{\sigma_{2}},
\end{equation*}
and:\\
\begin{align*}
&\hat{d}_{\theta_k}(s,a)_{2} = \hat{d}_{\theta_k}(s,a)_{3} = \dots = \hat{d}_{\theta_k}(s,a)_{|SA|} = 0, \\[8pt]
&\text{or} \\[8pt]
&\hat{d}_{\theta_k}(s,a)_{2} = \hat{d}_{\theta_k}(s,a)_{3} = \dots = \hat{d}_{\theta_k}(s,a)_{|SA|} \\[8pt]
&= \frac{2(\sigma_{2}-1)}{\sigma_{2}(|SA|-1)}.
\end{align*}
\end{lemma}

and finally, we have:

\begin{theorem}
The loosest acceptable $\sigma_{TV}$ covered at the $\sigma_2$ distance by our representative policy $\theta_{k}$:
\begin{align*}
& d_{\theta_{k}}(s,a)_{2} = d_{\theta_{k}}(s,a)_{3} = \dots = d_{\theta_{k}}(s,a)_{|SA|} \\[8pt]
& = \frac{\sigma_{2}-1}{\sigma_{2}(|SA|-1)}
\end{align*}
is given by:
\begin{equation*}
\boldsymbol{\sigma_{TV} = \frac{(\sigma_{2}-1)}{\sigma_{2}}}.
\end{equation*}
\end{theorem}

We can also find a policy that is at a $\sigma_{2}$ distance and simultaneously at the shortest Euclidean distance from our representative policy $\theta_{k}$.

To identify such a policy, we can use the following lemma:

\begin{lemma}
By moving away from the representative policy $\theta_{k}$ with fixed $\hat{d}_{\theta_k}(s,a)_{1}$ equal to $\frac{1}{\sigma_{2}}$, and in the direction of a specific vertex, for instance $(s,a)_{2}$, at a $\sigma_{2}$ distance away from the representative policy $\theta_{k}$:
\begin{align*}
& d_{\theta_{k}}(s,a)_{2} = d_{\theta_{k}}(s,a)_{3} = \dots = d_{\theta_{k}}(s,a)_{|SA|} \\[8pt]
& = \frac{\sigma_{2}-1}{\sigma_{2}(|SA|-1)}
\end{align*}
and considering:
\begin{align*}
&\hat{d}_{\theta_k}(s,a)_{3} = \hat{d}_{\theta_k}(s,a)_{4} = \dots = \hat{d}_{\theta_k}(s,a)_{|SA|} \\[8pt]
&= \frac{1 - \frac{1}{\sigma_{2}} - \hat{d}_{\theta_k}(s,a)_{2}}{|SA|-2},
\end{align*}
we reach:
\begin{equation*}
\hat{d}_{\theta_k}(s,a)_{2} = \frac{\frac{1}{|SA|-2} \pm \sqrt{\frac{\sigma_{2}}{|SA|-2}}}{\frac{\sigma_{2}(|SA|-1)}{(\sigma_{2}-1)(|SA|-2)}}.
\end{equation*}
and:\\
\begin{align*}
&\hat{d}_{\theta_k}(s,a)_{3} = \hat{d}_{\theta_k}(s,a)_{4} = \dots = \hat{d}_{\theta_k}(s,a)_{|SA|}\\[8pt]
&= \frac{1-\frac{1}{\sigma_{2}}-\left(\frac{\frac{1}{|SA|-2} \pm \sqrt{\frac{\sigma_{2}}{|SA|-2}}}{\frac{\sigma_{2}(|SA|-1)}{(\sigma_{2}-1)(|SA|-2)}}\right)}{|SA|-2}.
\end{align*}
\end{lemma}

and finally, we have:

\begin{theorem}
The tightest acceptable total variation $(\min\sigma_{TV})$ is the shortest radius of the ellipsoidal space covered at the $\sigma_2$ distance by our representative policy $\theta_{k}$:
\begin{align*}
& d_{\theta_{k}}(s,a)_{2} = d_{\theta_{k}}(s,a)_{3} = \dots = d_{\theta_{k}}(s,a)_{|SA|} \\[8pt]
& = \frac{\sigma_{2}-1}{\sigma_{2}(|SA|-1)}
\end{align*}
and is equal to:
\begin{equation*}
\boldsymbol{\min\sigma_{TV} = \frac{(\sigma_{2} - 1)\sqrt{(|SA| - 2)}}{\sqrt{\sigma_{2}}(|SA| - 1)}}.
\end{equation*}
\end{theorem}

Since we are seeking the upper bound, we can consider only the tightest distance $\min\sigma_{TV}$ to determine the space covered by the representative policy $\theta_{k}$.

Since we are in search of the upper bound, we can just take the shortest Euclidean distance $\min\sigma_{TV}$ into consideration in order to find the space covered by the representative policy $\theta_{k}$.

\subsection{Number of Samples with Known Model}

Now, if we transition from the $L_1$ norm metric to the Rényi divergence metric by examining the relationships of the tightest and loosest $\sigma_{TV}$ distances which are respectively $\min\sigma_{TV}$ and $\max\sigma_{TV}$ with the $\sigma_{2}$ distance within the polytope space, then the lower and upper bounds for the number of samples required for one policy are:
\begin{align}
 &N \geq \frac{k(2-\gamma_0)|SA|^2}{2\gamma_0(\sigma_{2}-1)(|SA|-1)}\ln\left(\frac{2}{\delta}\right)
 \nonumber\\[8pt]
 &N \leq \frac{k(2-\gamma_0)\sigma_{2}(|SA|-1)^2}{2\gamma_0(\sigma_{2}-1)^2(|SA|-2)}\ln\left(\frac{2}{\delta}\right).
 \nonumber
\end{align}

\subsection{Number of Samples with Unknown Model}

Similarly, for model-free cases, considering the tightest and loosest $\sigma_{TV}$ distances for the $\sigma_{2}$ distance in the Rényi divergence, the lower and upper bounds for the number of samples required are as follows:
\begin{align}
 &N \geq \frac{\gamma^{2}|S^{4}A|}{2(1-\gamma)^{2}(\sigma_{2}-1)(|S|-1)}\ln\left(\frac{2}{\delta}\right)
 \nonumber\\[8pt]
 &N \leq \frac{\gamma^{2}\sigma_{2}(|S|-1)^2|S^{2}A|}{2(1-\gamma)^{2}(\sigma_{2}-1)^{2}(|S|-2)}\ln\left(\frac{2}{\delta}\right).
 \nonumber
\end{align}

\section{Conclusion}

This paper addresses a critical challenge in reinforcement learning—sample inefficiency—by investigating the statistical foundations of policy space compression. By reducing the size of the policy space while preserving most of its decision-making capabilities, policy compression accelerates learning in environments with vast state-action spaces. Our research offers a detailed analysis of the sample complexity required to accurately learn this compressed policy set.

We achieve this by leveraging Rényi divergence to measure the discrepancy between the true and estimated policy distributions, ultimately deriving error bounds that ensure near-optimal policy performance with high confidence. Additionally, we simplify our analysis through the use of the $l_1$ norm, determining sample size requirements for both model-based and model-free settings. The correlation between the error bounds from the $l_1$ norm and those from Rényi divergence enables us to establish clear upper and lower bounds on the necessary sample sizes, distinguishing between policies near the vertices and those in the middle of the policy space.

Ultimately, our findings provide valuable insights into the sample efficiency of policy compression, offering a framework that balances learning speed and decision accuracy. This work lays the groundwork for future research in enhancing reinforcement learning algorithms by addressing the fundamental issue of sample complexity through the lens of policy space compression.

\end{document}